%% file: main.tex
\documentclass[10pt,twocolumn,letterpaper]{article}

\input{preamble}
\usepackage[pagenumbers]{cvpr}

\definecolor{cvprblue}{rgb}{0.21,0.49,0.74}
\usepackage[pagebackref,breaklinks,colorlinks,allcolors=cvprblue]{hyperref}
\usepackage{enumitem}

\title{\ours: Human LiDAR Generation from Parametric Body Models}

\author{
Salma Galaaoui\textsuperscript{1,2}\quad\;
Nermin Samet\textsuperscript{1}{$^*$}\quad\; 
David Picard\textsuperscript{2}{$^*$}\quad\; 
\\
\vspace{-2mm}
{
\textsuperscript{1} Valeo.ai, Paris, France
}
\and
{
\textsuperscript{2} LIGM, CNRS, Univ Gustave Eiffel, ENPC, Institut Polytechnique de Paris, Marne-la-Vallée, France
}
}

\begin{document}

\twocolumn[{%
  \renewcommand\twocolumn[1][]{#1}%
  \maketitle
  \begin{center}
    \captionsetup{type=figure}
    \input{figures/teaser}
  \end{center}
}]

\blfootnote{$^*$Equal contribution for senior authorship.}

\input{sec/0_abstract}    
\input{sec/1_intro}
\input{sec/2_rw}
\input{sec/3_method}
\input{sec/4_exp}
\input{sec/5_conclusion}
\clearpage

{
    \small
    \bibliographystyle{ieeenat_fullname}
    \bibliography{main}
}

\input{sec/X_suppl}

\end{document}

%% file: preamble.tex
\usepackage{epsfig}
\usepackage{graphicx}
\usepackage{amsmath}
\usepackage{amssymb,mathtools}
\usepackage{siunitx}
\newrobustcmd{\B}{\bfseries}

\usepackage{etoolbox}

\usepackage{bbding}
\usepackage{bm}
\usepackage{booktabs}
\usepackage{dblfloatfix}
\usepackage[accsupp]{axessibility}  
\usepackage{xspace}
\usepackage[dvipsnames]{xcolor}
\usepackage{comment}
\usepackage{array}
\usepackage{makecell}
\usepackage{pifont}
\usepackage{multirow}
\usepackage{flushend}
\usepackage{colortbl}
\usepackage{comment} 
\usepackage{sidecap} 
\usepackage{nicefrac} 
\newcolumntype{R}[2]{%
    >{\adjustbox{angle=#1,lap=1.3\width-(#2)}\bgroup}%
    l%
    <{\egroup}%
} 
\usepackage{pgfkeys}
\usepackage[]{pgfplots,pgfplotstable}
\usepackage{algorithm}
\usepackage{algpseudocode}

\usepackage[export]{adjustbox} 

\usepackage[]{tikz}
\usetikzlibrary{arrows, positioning, math, fit, backgrounds}
\usepackage{numprint,fullpage}  
\usepackage[nomessages]{fp}

\usepackage{cancel}

\pgfplotsset{compat=1.11,
    /pgfplots/ybar legend/.style={
    /pgfplots/legend image code/.code={%
       \draw[##1,/tikz/.cd,yshift=-0.25em]
        (0cm,0cm) rectangle (3pt,0.8em);},
   },
}

\makeatletter
\renewcommand\paragraph{\@startsection{paragraph}{4}{\z@}%
    {0.8ex \@plus 1ex \@minus .1ex}%
    {-1em}%
    {\normalfont \normalsize \bfseries}}
\makeatother

\def\addlegendimage{\csname pgfplots@addlegendimage\endcsname}

\newcommand\ours{\textsl{HuLiGen}\xspace}

\newcommand\sloper{SLOPER4D\xspace}
\newcommand\waymo{WOD\xspace}

\definecolor{myblue}{rgb}{0.19, 0.55, 0.91}
\definecolor{myred}{rgb}{0.82, 0.1, 0.26}
\definecolor{MyGreen}{RGB}{0, 104, 55} 
\definecolor{MyRed}{RGB}{248, 3, 7} 
\definecolor{MyYellow}{RGB}{180, 180, 0} 
\newcommand{\cmark}{{\textcolor{MyGreen}{\ding{51}}}}%
\newcommand{\xmark}{{\textcolor{MyRed}{\ding{55}}}}%
\newcommand\rf[2]{\,{\scriptsize #1\,'#2}}
\newcommand\NA{\mbox{\scriptsize\color{gray} N/A}}
\newcommand\na{\mbox{\scriptsize\color{gray} N/A}}

\newlength \mywidth
\newcommand\blfootnote[1]{%
  \begingroup
  \renewcommand\thefootnote{}\footnote{#1}%
  \addtocounter{footnote}{-1}%
  \endgroup
}
\usepackage{tikz}
\usetikzlibrary{positioning,arrows.meta,backgrounds,calc,fit}
\usepackage{fontawesome5}

%% file: figures/teaser.tex
\providecommand{\qbase}{figures/teaser_viridis}

\ifdefined\qh\else\newlength{\qh}\fi
\setlength{\qh}{2.20cm}

\ifdefined\qw\else\newlength{\qw}\fi
\setlength{\qw}{0.72\qh}

\ifdefined\qv\else\newlength{\qv}\fi
\setlength{\qv}{4pt}

\providecommand{\qc}[1]{%
  \makebox[\qw][c]{%
    \includegraphics[height=\qh]{\qbase/#1}%
  }%
}

\providecommand{\qhd}[1]{%
  \makebox[\qw][c]{\footnotesize #1}%
}

\providecommand{\qn}[1]{%
  \makebox[\qw][c]{\scriptsize #1}%
}

\providecommand{\qlab}[1]{%
  \rotatebox[origin=c]{90}{\footnotesize\textsc{#1}}%
}

\providecommand{\qpair}[8]{%
  %
  \qc{#1/rgb.png} &
  \qc{#1/gt_mesh.pdf} &
  \qc{#1/gt_pc_#2.pdf} &
  \qc{#1/ray_cast_#3.pdf} &
  \qc{#1/ours_#4.pdf} &
  %
  \qc{#5/rgb.png} &
  \qc{#5/gt_mesh.pdf} &
  \qc{#5/gt_pc_#6.pdf} &
  \qc{#5/ray_cast_#7.pdf} &
  \qc{#5/ours_#8.pdf}
  \tabularnewline[1pt]
  %
    \multicolumn{2}{r}{\scriptsize \# pts\hspace{3pt}} &
    \qn{#2} &
    \qn{#3} &
    \qn{#4} &
    \multicolumn{2}{r}{\scriptsize \# pts\hspace{3pt}} &
    \qn{#6} &
    \qn{#7} &
    \qn{#8}
  \tabularnewline[\qv]
}

\centering

\setlength{\tabcolsep}{0pt}
\renewcommand{\arraystretch}{0}

\begin{tabular}{@{}c@{\hspace{2pt}}c@{}}

&
\begin{tabular}{
  @{}
  ccccc
  @{\hspace{3pt}}
  ccccc
  @{}
}
  \qhd{RGB} &
  \qhd{GT mesh} &
  \qhd{GT PC} &
  \qhd{Raycasted} &
  \qhd{\ours} &
  \qhd{RGB} &
  \qhd{GT mesh} &
  \qhd{GT PC} &
  \qhd{Raycasted} &
  \qhd{\ours}
\end{tabular}
\tabularnewline[3pt]

 &
\begin{tabular}{
  @{}
  cc
  c
  cc
  @{\hspace{10pt}}
  cc
  c
  cc
  @{}
}

  \qpair
    {waymo_val_30779396576054160_1880_000_1900_000_TSKM92rlKS_KCqprYcZPyA_x-1557845090164314}{1032}{542}{1032}
{sloper4d_train_seq005_library_002-529_707}{2486}{2058}{2486}

\end{tabular}

\end{tabular}


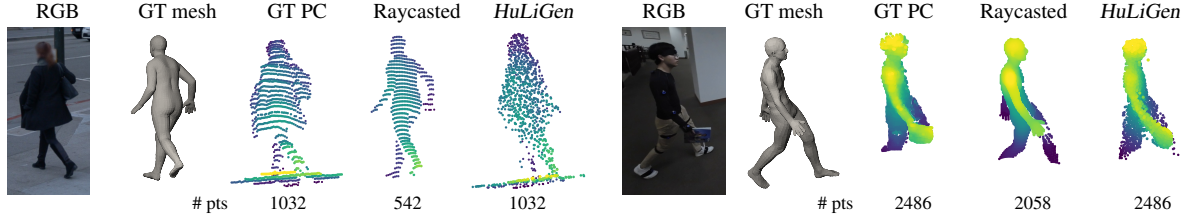
\captionof{figure}{\textbf{Learning realistic human LiDAR observations from real data.}
Existing synthetic pipelines generate articulated 
human point clouds by ray casting parametric body models, which represent the underlying body surface but omit clothing, hair and other surface variations.
This also affects the resulting point distribution, often leading to fewer and less consistent points than in real observations. In the left example, the woman’s coat is not rendered and sensor-specific noise and measurement irregularities are largely absent from the simulated sample.
On the right, the book is captured by our generation while absent in ray casting.
In contrast, our method learns these characteristics directly from real data, producing point clouds with more realistic geometry, point distributions and sensor effects.
The camera images are shown only for visualization and are not used by our pipeline.
}
\label{fig:teaser}

%% file: sec/0_abstract.tex
\begin{abstract}
LiDAR point clouds of humans are extremely expensive to collect and annotate, thus represent a scarce resource that hinders the development of human analysis using this modality. 
To alleviate this scarcity, prior work relies on simulated human LiDAR, but such samples do not fully reflect the geometry and sensing characteristics of real observations.
In contrast, we introduce \ours, a generative model that generates human LiDAR point clouds from a parametric body model, using a point transformer trained with a flow-matching objective. We show that our generated point clouds are closer to the real capture distribution. 
Using \ours to generate synthetic data, we propose a synthetic-only pretraining scheme for LiDAR-based HPE that achieves state-of-the-art performance, with even larger gains in low-annotation and low-data regimes, where MPJPE is reduced by up to 50\%. Code, models and generated samples are available at \url{https://github.com/valeoai/HuLiGen}.
\end{abstract}

%% file: sec/1_intro.tex
\section{Introduction}
\label{sec:intro}

3D human pose estimation (HPE) is an important task for human-centric perception in embodied systems. Accurate estimation of human body pose can provide fine-grained cues for pedestrian behavior understanding and intention prediction \cite{razali2021pedestrian}, as well as trajectory forecasting~\cite{tripod,jeong2025trajectory}, which are important for collision avoidance and interaction-aware motion planning~\cite{neumann2021pedestrian}. 
The geometric nature of LiDAR makes it particularly attractive for these applications, but also introduces a fundamental bottleneck: human LiDAR data with accurate 3D pose annotations remains difficult to obtain at scale.

Constructing such datasets requires substantial annotation effort. LiDAR points on humans are sparse, viewpoint dependent, and frequently affected by self-occlusion and external occlusions, while accurate pose supervision typically requires synchronized motion-capture systems, additional cameras or inertial sensors.
Recent datasets such as SLOPER4D~\cite{sloper4d} and the Waymo Open Dataset (WOD)~\cite{waymo} provide real-world point clouds together with human pose annotations which  help the progress in LiDAR-based human pose estimation.
Nevertheless, the available data still covers only a limited range of human poses and body shapes.
For example, SLOPER4D contains more than 100K LiDAR frames but only 12 recorded subjects, while WOD provides approximately 10K object frames with 3D keypoint annotations. 
In contrast, large-scale human body-model repositories such as AMASS~\cite{amass}, provide a much richer diversity of human poses and body configurations, but lack corresponding LiDAR observations.

To compensate for the limited amount and diversity of annotated human LiDAR data, a common strategy is to generate synthetic samples from parametric body models using ray casting~\cite{dapt,gckpl,livehpslus,unipvu}. However, the realism of these samples is limited by the manually defined rendering process~\cite{lidarsim,lidargen}. 
Real LiDAR observations depend not only on the underlying body geometry, but also on clothing and surface variations, viewpoint, range, sensor configuration and measurement uncertainty.
Modeling these effects requires explicit assumptions and hand-crafted noise or sensor models.

In this work, we explore a different paradigm: rather than explicitly simulating the interaction of the LiDAR with the body model to obtain the point cloud, we learn it directly from real lidar observations. Given real human point clouds paired with corresponding body models, we train a conditional flow-matching model that acts as a learned LiDAR renderer, modeling how a given human pose and shape is observed by a real sensor. This formulation learns real LiDAR characteristics such as sparsity, viewpoint- and range-dependent sampling and uncertainty. It can also capture deviations from the underlying body model caused by clothing, hair and other surface variations (see \cref{fig:teaser}). Finally, it decouples pose diversity from sensor realism: large-scale mocap data provides diverse human poses, while real LiDAR data defines how these poses are observed.

Once trained, the model can generate LiDAR observations for previously unseen poses, allowing us to leverage existing large-scale mocap datasets with diverse poses and body configurations to generate sensor-specific human point clouds.
Beyond evaluating generation quality using standard generative metrics, we further assess the generated point clouds on the downstream task of 3D human pose estimation. We show that pretraining a pose estimator on generated human LiDAR samples, followed by finetuning on real data, significantly improves performance in both full-data and low-data regimes. Experiments with different fractions of the real training set further show that our generated data can compensate for limited labeled real data.

Our main contributions are:

\begin{itemize}[nosep]
    \item We introduce \ours, to the best of our knowledge, the first learned generative mapping from parametric human body models to LiDAR point-clouds, learning real LiDAR characteristics.
    \item We demonstrate the practical value of the learned LiDAR distribution through downstream 3D HPE. Pretraining on generated human LiDAR observations followed by finetuning on real data reduces MPJPE from 6.02 cm to 5.04 cm on SLOPER4D and from 6.74 cm to 5.85 cm on WOD, achieving state-of-the-art performance on both datasets.
    \item We demonstrate strong gains in the low-annotation setting on \waymo, where our pretraining pipeline reduces MPJPE by about 50\% using only 1\% and 10\% of the available pose annotations.
    \item We further show that our pretraining pipeline remains effective in low-data settings, reducing MPJPE by about 30\% and 25\% on SLOPER4D \cite{sloper4d}, and by 42\% and 36\% on WOD \cite{waymo}, when using only 10\% and 20\% of the real training data, respectively.
     
\end{itemize}

%% file: sec/2_rw.tex
\section{Related works}
\label{sec:rw}

\paragraph{Lidar-based 3D human pose estimation.}
Lidar-based human pose estimation is challenging due to the sparsity, non-uniform density and partial observation of the human body. Early approaches therefore focused on extracting more structured representations from sparse measurements. LidPose~\cite{kovacsLidPoseRealTime3D2024} projects human point clouds into structured 2D representations, LPFormer~\cite{yeLPFormerLiDARPose2024} combines BEV and sparse voxel features, and VoxelKP~\cite{shiVoxelKPVoxelbasedNetwork2023} operates directly on high-resolution sparse voxels. 
LiDAR-HMR~\cite{prn} leverages Pose Regression Network (PRN) to estimate 3D keypoints directly from human LiDAR point clouds through probabilistic point-wise voting and attention-based refinement.
Other works compensate for missing geometric information using complementary cues: MMVP~\cite{fanHumanM3MultiviewMultimodal2023} and HUM3DIL~\cite{zanfirHUM3DILSemisupervisedMultimodal2022} fuse RGB and LiDAR information, while LiDAR-HMP~\cite{hanPracticalHumanMotion2024} exploits temporal context. More recently, the focus has shifted toward learning stronger human-specific representations. UniPVU-Human~\cite{unipvu} learns body-part and motion priors before downstream pose estimation, while DAPT~\cite{dapt} explicitly addresses variations in LiDAR point density through density-aware feature learning. Beyond architectural improvements, GC-KPL~\cite{gckpl} tackles the limited availability of labeled LiDAR data by combining synthetic pretraining with geometric and temporal self-supervision on unlabeled real data.

\paragraph{Synthetic data for lidar-based 3D HPE.}
Beyond developing better architectures and representations, prior work has also attempted to address the scarcity of annotated human LiDAR data. One line of research introduces additional supervision from images, using 2D annotations or pseudo-labels together with LiDAR to reduce the dependence on 3D pose labels~\cite{furstHPERL3DHuman2021,zhengMultimodal3DHuman2022,bauerWeaklySupervisedMultiModal2023,congWeaklySupervised3D2023,panLiCamPoseCombiningMultiView2025}. For LiDAR-only methods, two strategies are commonly adopted to increase training diversity and quantity. VoxelKP~\cite{shiVoxelKPVoxelbasedNetwork2023}, LPFormer~\cite{yeLPFormerLiDARPose2024}, and LidPose~\cite{kovacsLidPoseRealTime3D2024} rely on extensive data augmentation, such as frustum dropout, frame mixing, body cropping and geometric transformations, to perturb existing real samples. 
A second direction generates entirely new LiDAR observations from parametric human models using simulated LiDAR ray casting and hand-designed perturbations, such as laser-level masks for occlusion~\cite{dapt,gckpl,livehpslus,unipvu}.
This makes it possible to exploit large-scale body-model repositories such as AMASS~\cite{amass}, even though they do not contain paired LiDAR measurements. In contrast, our approach preserves this ability to leverage large-scale human body data while learning the mapping from body models to LiDAR observations directly from real data.

\noindent{\textbf{Lidar point cloud generation.}}
LiDAR generation has evolved from raycasting simulators in virtual environments~\cite{fang2020augmentedLiDAR} to learned generative models based on energy functions, discrete latents and diffusion~\cite{lidargen,xiong2023ultraLiDAR,nakashima2023r2dm,ran2024lidm}. Most of these methods, however, focus on full-scene generation rather than the fidelity of individual objects.
Recent object-level LiDAR generators explicitly model foreground objects. These methods generate localized objects with diffusion, using conditioning such as object class, position, or viewpoint, and can integrate the generated instances back into complete driving scenes~\cite{yan2025olidm,logen}.
Despite this progress, existing LiDAR generators are not designed to generate humans conditioned on their underlying body pose and shape.

%% file: sec/3_method.tex
\section{Method}
\label{sec}
\subsection{Overview}
\input{figures/canonicalization/root_axes_panels2}
\input{figures/pipeline}

Our goal is to learn to predict how a LiDAR would capture a human, as represented by a parametric body model. Formally, given a human point cloud $\mathbf{P}$ and its corresponding body model $\mathbf{B}$, we want to learn the conditional distribution
\begin{align}
    p(\mathbf{P} \mid \mathbf{B}),
\end{align}
where $\mathbf{B}$ can be further decomposed into $[\mathbf{o},\beta, \theta]$ the oriented location of the subject, the body shape and local body pose representations respectively.

We model this conditional distribution using flow matching (FM) parametrized by a neural network operating directly on the 3D point cloud. Starting from a random Gaussian point cloud, this flow-matching model learns to gradually move the 3D points towards that of a human LiDAR capture, conditionally to $[\mathbf{o}, \beta, \theta]$ that correspond to the body model $\mathbf{B}$ of the human. Because each LiDAR produces very different distributions of point clouds, we train a separate model for each acquisition setting so that sensor-dependent characteristics are captured independently.

\subsection{Human-centric parameterization}
Because humans are articulated and their capture by a LiDAR results in very local interactions, the choice of coordinate system is crucial to provide relevant information to the flow-matching model. Let us denote by subscript $w$ the world coordinates and by subscript $l$ the LiDAR coordinates.

The human point cloud $\mathbf{P} = \{ \mathbf{p}_i\}_{1\leq i\leq N}$ is usually not touching the surface of the body model because of clothing and noise. Since any side of the human can be captured, the distribution of $\mathbf{P}$ is centered around the body model in expectation over all possible captures. As such, we propose to use the center of the body model in LiDAR coordinates as the origin. Noting $\mathbf{t}_w$ the center of the body model in world coordinates, its LiDAR counterpart is given by using a translation $\mathbf{t}_l=\mathbf{T}_{w\rightarrow l}\mathbf{t}_w$, and each point of $\mathbf{P}$ is expressed relative to it:
\begin{align}
    x_i = \mathbf{T}_{w\rightarrow l}\mathbf{p}_i - \mathbf{t}_l.
\end{align}

To encode the distance and orientation of the body with respect to the sensor (the $\mathbf{t}_l$ part of $\mathbf{B}$), we use cylindrical coordinates:
\begin{align}
\phi=\operatorname{atan2}(t_{l_y},t_{l_x}),\quad \rho=\sqrt{t_{l_x}^2+t_{l_y}^2},\quad z=t_{l_z}.
\end{align}
$\phi$, $\rho$, and $z$ respectively describe azimuth, radial distance and height relative to the LiDAR.

In addition to center on the body, we also adopt a cannonical reference frame aligned on the global body yaw. The reasoning behind this step is to let the flow-matching model operate in a frame that does not depend on the pitch bend in the pose (\textit{e.g.}, a person tying their shoes) as humans tend to not tilt side to side (see \cref{fig:root_axes_poses2}). More formally, denoting $\mathbf{R}_0$ the root joint rotation matrix, the azimuth of the lateral axis is obtained with
\begin{align}
\psi =
\operatorname{atan2}
\left(
[\mathbf{R}_0]_{10},
[\mathbf{R}_0]_{00}
\right).  
\end{align}
Then, the points are transformed into the canonical body frame with the rotation
\begin{align}
\resizebox{0.88\linewidth}{!}{$
\mathbf{R}_z(-\psi)\cdot\mathbf{x}_i, \quad
\mathbf{R}_z(-\psi) =
\begin{bmatrix}
\cos\psi & \sin\psi & 0\\
-\sin\psi & \cos\psi & 0\\
0 & 0 & 1
\end{bmatrix}
$}
\end{align}
and the same rotation is applied to all body joints to keep a consistent reference frame. 
Similarly, we apply this correction to the azimuth encoding of the LiDAR by computing $\phi_{\mathrm{rel}} = ((\phi - \psi + \pi)\bmod 2\pi)-\pi$.
After the generation, the inverse rotation is applied to recover the generated point cloud in the original LiDAR frame.

\subsection{Condition encoding}

Our conditioning $\mathbf{B}$ combines three complementary sources of information:
\begin{equation}
\Big(
\underbrace{\phi_{\mathrm{rel}}, \rho, z}_{\text{oriented body location }\mathbf{o}}, \,
\underbrace{\beta\vphantom{\phi_{\mathrm{rel}}}}_{\text{body shape}}, \,
\underbrace{\theta\vphantom{\phi_{\mathrm{rel}}}}_{\text{local body pose}}
\Big),
\end{equation}
that require specific encoding each.

\paragraph{Oriented body location condition.}
Because angular variables are discontinuous at $-\pi$ and $\pi$, we first represent $\phi_{\mathrm{rel}}$ through its sine and cosine components and apply a cyclic embedding. The radial distance $\rho$ and height $z$ are represented using Fourier features. This provides 3 sensor-related condition tokens.

\paragraph{Body shape condition.}
Using the parametric body model for the shape, each of the shape coefficients is independently embedded into a different token using an MLP to keep these parameters disentangled, leading to different condition tokens.

\paragraph{Local body pose condition.} 
Using the body model for the pose $\theta$, we convert the axis-angle to 3D cartesian coordinates using forward kinematics and embed each of the 3D body joints using a PointNet~\cite{qi2017pointnet} model to obtain one token per joint. To retain the semantic information of the joints, we keep their order the same and add a learned positional encoding.

The condition encoding results in the concatenation of all the tokens in a single sequence of tokens.

\subsection{3D Point cloud flow transformer}
To generate the 3D LiDAR point cloud from the condition embedding tokens, we train a point-based Diffusion Transformer~\cite{dit3d} with a flow-matching objective~\cite{lipmanflow}.

Given a human point cloud in canonical frame $\mathbf{X}_1 \in \mathbb{R}^{N\times 3}$ and a random initial point cloud $\mathbf{X}_0 \sim \mathcal{N}(0, 1)$, we define the linear interpolant $\mathbf{X}_t = t\mathbf{X}_1 + (1-t)\mathbf{X}_0, t\sim \mathcal{U}(0,1)$, that corresponds to the following differential equation associated with the velocity field $v_t$:
{
\begin{align}
    \mathrm{d}\mathbf{X}_t = v_t \mathrm{d}t,\\
    v_t = \mathbf{X}_1 - \mathbf{X}_0.
\end{align}
}
Following this velocity field allows to recover $\mathbf{X}_1$ starting from the random point cloud $\mathbf{X}_0$.

The transformer is trained to approximate $v_t$. It takes as input $\mathbf{X}_t, t$ and the condition embedding tokens, and output a velocity field $v_\Theta(\mathbf{X}_t, t, \mathbf{B})$, optimized using the following regression loss:
\begin{align}
    \mathcal{L} = \| v_t - v_\Theta(\mathbf{X}_t, t, \mathbf{B}) \|^2,
\end{align}
optimized by back-propagation.

Once the transformer is trained, we generate novel LiDAR point clouds by simply sampling an initial random point cloud $\mathbf{X}_0\sim \mathcal{N}(0,1)$ and solving the differential equation $\mathbf{X}_t = v_\Theta(\mathbf{X}_t, t, \mathbf{B}) \mathrm{d}t$ using an Euler sampling scheme. Notice that using a transformer based architecture allows us to sample varying amounts of points $N$ without the need for architectural changes.

%% file: figures/canonicalization/root_axes_panels2.tex
\providecommand{\rapath}{figures/canonicalization}
\newlength{\rah}\setlength{\rah}{2.0cm}

\begin{figure}[t]\centering
\definecolor{axlat}{HTML}{2E7D4F}   
\definecolor{axfwd}{HTML}{B4322A}   
\definecolor{axup}{HTML}{8895A4}    
  \begin{subfigure}[b]{0.2\textwidth}\centering
    \includegraphics[height=\rah]{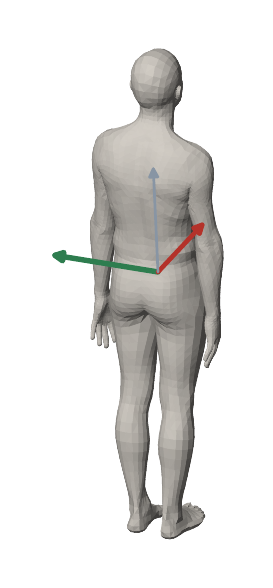}
    \caption{standing}
  \end{subfigure}\hfill
  \begin{subfigure}[b]{0.2\textwidth}\centering
    \includegraphics[height=\rah]{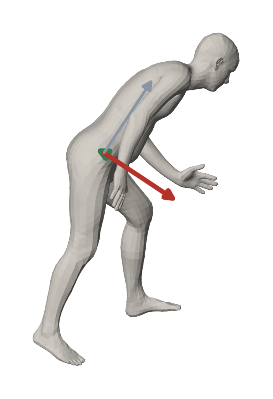}
    \caption{bent forward}
  \end{subfigure}\hfill
  \\[8pt]
  \begin{subfigure}[b]{0.2\textwidth}\centering
    \includegraphics[height=\rah]{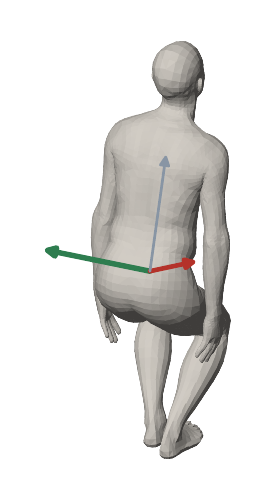}
    \caption{squatting}
  \end{subfigure}\hfill
  \begin{subfigure}[b]{0.2\textwidth}\centering
    \includegraphics[height=\rah]{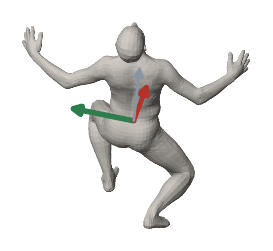}
    \caption{crouched}
  \end{subfigure}\hfill
  \\[8pt]
  \begin{subfigure}[b]{0.2\textwidth}\centering
    \includegraphics[height=\rah]{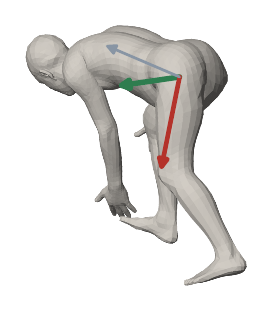}
    \caption{forward near vertical}
  \end{subfigure}\hfill
  \begin{subfigure}[b]{0.2\textwidth}\centering
    \includegraphics[height=\rah]{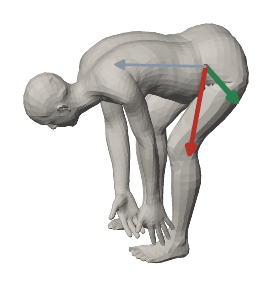}
    \caption{forward near vertical}
  \end{subfigure}\hfill
  \\[8pt]
  \begin{subfigure}[b]{0.2\textwidth}\centering
    \includegraphics[height=\rah]{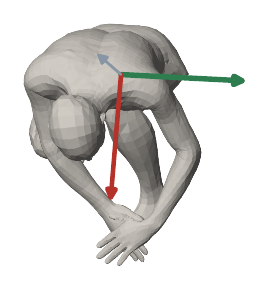}
    \caption{forward near vertical}
  \end{subfigure}\hfill
  \begin{subfigure}[b]{0.2\textwidth}\centering
    \includegraphics[height=\rah]{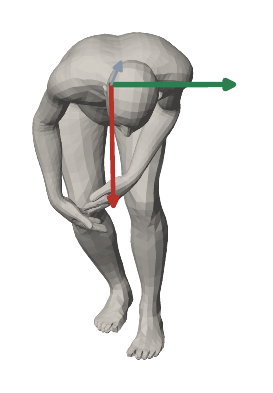}
    \caption{forward near vertical}
  \end{subfigure}\hfill
  \caption{
  \textbf{Root-joint axes across different pose types from the \sloper~\cite{sloper4d} dataset.}
We use the azimuth of the \textcolor{axlat}{\textbf{\emph{lateral}}} axis for canonicalisation, as its projection onto the horizontal plane remains well defined across the different poses. In contrast, the \textcolor{axfwd}{\textbf{\emph{forward}}} axis can approach the vertical under strong hip flexion, causing its horizontal projection to shrink (at the risk of reaching 0$^\circ$) and making the corresponding azimuth ill-conditioned.
  }
  
  \label{fig:root_axes_poses2}
\end{figure}

%% file: figures/pipeline.tex
\providecommand{\pipebase}{figures/teaser_viridis}

\providecommand{\imgMesh}   {amass_train_314-walking_slow08_stageii-79/gt_mesh.pdf}
\providecommand{\imgSynth}  {amass_train_314-walking_slow08_stageii-79/ours_2275.pdf}
\providecommand{\imgSynthKp}{amass_train_314-walking_slow08_stageii-79/keypoints_gt.pdf}
\providecommand{\imgReal}   {sloper4d_val_seq009_running_002-1218_004/gt_pc_2081.pdf}
\providecommand{\imgRealKp} {sloper4d_val_seq009_running_002-1218_004/keypoints_gt.pdf}

\providecommand{\pipenewlen}[1]{\ifdefined#1\relax\else\newlength{#1}\fi}
\providecommand{\PL}[1]{\the\dimexpr#1\relax}  

\pipenewlen{\pipefigw}  \setlength{\pipefigw} {\textwidth} 
\pipenewlen{\pipeh}     \setlength{\pipeh}    {1.60cm}     
\pipenewlen{\pipegap}   \setlength{\pipegap}  {5mm}        
\pipenewlen{\pipedbx}   \setlength{\pipedbx}  {5mm}        
\pipenewlen{\pipedby}   \setlength{\pipedby}  {6.5mm}      
\pipenewlen{\pipepad}   \setlength{\pipepad}  {4mm}        
\pipenewlen{\pipepadb}  \setlength{\pipepadb} {5.5mm}      
\pipenewlen{\pipegapx}  \setlength{\pipegapx} {3mm}        
\pipenewlen{\pboxtw}    \setlength{\pboxtw}   {1.75cm}     

\providecommand{\pipeimg}[1]{\includegraphics[height=\pipeh]{\pipebase/#1}}

\pipenewlen{\pboxw}
\setlength{\pboxw}{\pboxtw}\addtolength{\pboxw}{3pt}   

\providecommand{\pipemeas}[2]{\settowidth{#1}{\includegraphics[height=\pipeh]{\pipebase/#2}}}

\pipenewlen{\wgen}\pipenewlen{\wkpi}\pipenewlen{\wreal}\pipenewlen{\wkpii}
\pipemeas{\wgen} {\imgSynth}
\pipemeas{\wkpi} {\imgSynthKp}
\pipemeas{\wreal}{\imgReal}
\pipemeas{\wkpii}{\imgRealKp}

\pipenewlen{\sumI}\pipenewlen{\sumII}\pipenewlen{\pipetmp}
\setlength{\sumI} {\wgen} \addtolength{\sumI} {\wkpi}
\setlength{\sumII}{\wreal}\addtolength{\sumII}{\wkpii}

\pipenewlen{\padI}\pipenewlen{\padII}
\setlength{\padI}{\pipepad}\setlength{\padII}{\pipepad}
\ifdim\sumI<\sumII
  \setlength{\pipetmp}{\sumII}\addtolength{\pipetmp}{-\sumI}%
  \addtolength{\padI}{0.5\pipetmp}%
\else
  \setlength{\pipetmp}{\sumI}\addtolength{\pipetmp}{-\sumII}%
  \addtolength{\padII}{0.5\pipetmp}%
\fi

\pipenewlen{\panelw}
\setlength{\panelw}{\sumI}
\addtolength{\panelw}{2\pipegap}
\addtolength{\panelw}{\pboxw}
\addtolength{\panelw}{2\padI}

\pipenewlen{\ttlw}
\settowidth{\ttlw}{\normalsize\bfseries Pretraining on synthetic data}
\settowidth{\pipetmp}{\normalsize\bfseries Finetuning on real data}
\ifdim\pipetmp>\ttlw \setlength{\ttlw}{\pipetmp}\fi
\addtolength{\ttlw}{4mm}
\ifdim\panelw<\ttlw
  \setlength{\pipetmp}{\ttlw}\addtolength{\pipetmp}{-\panelw}%
  \addtolength{\padI} {0.5\pipetmp}%
  \addtolength{\padII}{0.5\pipetmp}%
  \setlength{\panelw}{\ttlw}%
\fi

\pipenewlen{\gapGtoI}\pipenewlen{\gapItoII}
\setlength{\gapGtoI}{\pipepad}\addtolength{\gapGtoI}{\pipegapx}\addtolength{\gapGtoI}{\padI}
\setlength{\gapItoII}{\padI}  \addtolength{\gapItoII}{\pipegapx}\addtolength{\gapItoII}{\padII}

\definecolor{pipeGen}{HTML}{000000}
\definecolor{pipeReal}{HTML}{000000}
\definecolor{pipeBox}{HTML}{243B53}
\definecolor{pipeRule}{HTML}{9AA5B1}
\definecolor{pipeDb}{HTML}{243B53}

\tikzset{
  dbicon/.pic={
    \draw[pipeDb,line width=0.8pt] (-3.4mm, 2.9mm) -- (-3.4mm,-2.9mm);
    \draw[pipeDb,line width=0.8pt] ( 3.4mm, 2.9mm) -- ( 3.4mm,-2.9mm);
    \draw[pipeDb,line width=0.8pt]
      (0,2.9mm) ellipse [x radius=3.4mm, y radius=1.25mm];
    \draw[pipeDb,line width=0.8pt]
      (-3.4mm,0mm)
      arc[start angle=180, end angle=360, x radius=3.4mm, y radius=1.25mm];
    \draw[pipeDb,line width=0.8pt]
      (-3.4mm,-2.9mm)
      arc[start angle=180, end angle=360, x radius=3.4mm, y radius=1.25mm];
  }
}

\begin{figure*}[t]
\centering

\resizebox{\pipefigw}{!}{%
\begin{tikzpicture}[
font=\small,
dbnode/.style={
  inner sep=0pt, outer sep=0pt,
  minimum width=6.8mm, minimum height=8.4mm
},
pbox/.style={
  draw=pipeBox,
  line width=0.9pt,
  rounded corners=2pt,
  text width=\pboxtw,
  inner xsep=1.5pt,
  outer sep=0pt,
  minimum height=1.08cm,
  align=center,
  fill=pipeBox!5
},
pimg/.style={inner sep=0pt, outer sep=0pt},
plbl/.style={font=\footnotesize},
pcap/.style={font=\footnotesize\itshape, text=black!65},
pstage/.style={font=\footnotesize\bfseries, text=black!72, align=center},
parrow/.style={
  -{Stealth[length=2.4mm,width=2.0mm]},
  line width=1.15pt, draw=pipeBox!85, rounded corners=3pt
},
pdash/.style={
  -{Stealth[length=2.4mm,width=2.0mm]},
  line width=1.0pt, draw=pipeBox!55,
  dashed, dash pattern=on 3pt off 2.2pt, rounded corners=3pt
},
]

\node[dbnode] (mocapdb) at (0,0) {};
\pic at (mocapdb.center) {dbicon};
\node[plbl,below=0.6mm of mocapdb] (mocapdblab) {Mocap DB};

\node[dbnode,anchor=south] (targetdb)
  at ($(mocapdb.north)+(0,\PL{\pipedby})$) {};
\pic at (targetdb.center) {dbicon};
\node[plbl,below=0.6mm of targetdb] (targetdblab) {Target dataset};

\coordinate (boxTop) at ($(targetdb.north)+(0,2.5mm)$);

\node[pimg,anchor=west] (mesh)
  at ($(mocapdb.east)+(\PL{\pipedbx},0)$) {\pipeimg{\imgMesh}};
\node[plbl,below=0.5mm of mesh] (meshlab) {local body pose $\boldsymbol{\theta}$};

\draw[parrow] (mocapdb.east) -- (mesh.west);

\node[pbox,anchor=west] (fm) at ($(mesh.east)+(\PL{\pipegap},0)$)
  {FM model\\[2pt]
   {\scriptsize\textcolor{pipeBox}{\faSnowflake}\; frozen}};

\draw[parrow] (mesh.east) -- (fm.west);

\coordinate (condturn) at (fm.north |- targetdb);
\draw[parrow] (targetdb.east)
  -- node[plbl,above=0.8mm,fill=white,inner sep=1pt]
       {$\# pts,\ \boldsymbol{o},\ \boldsymbol{\beta}$}
  (condturn) -- (fm.north);

\node[pimg,anchor=west] (gen) at ($(fm.east)+(\PL{\gapGtoI},0)$)
  {\pipeimg{\imgSynth}};
\node[plbl,above=0.5mm of gen] (genlab) {Synt. dataset};

\draw[parrow] (fm.east) -- (gen.west);

\node[pbox,anchor=west] (hpe1) at ($(gen.east)+(\PL{\pipegap},0)$)
  {HPE model\\[2pt]
   {\scriptsize\textcolor{pipeGen}{\faFire}\; trained}};

\node[pimg,anchor=west] (kp1) at ($(hpe1.east)+(\PL{\pipegap},0)$)
  {\pipeimg{\imgSynthKp}};
\node[plbl,above=0.5mm of kp1] (kp1lab) {3D joints};

\draw[parrow] (gen.east)  -- (hpe1.west);
\draw[parrow] (hpe1.east) -- (kp1.west);

\node[pimg,anchor=west] (gtpc) at ($(kp1.east)+(\PL{\gapItoII},0)$)
  {\pipeimg{\imgReal}};
\node[plbl,above=0.5mm of gtpc] (gtlab) {Target dataset};

\node[pbox,anchor=west] (hpe2) at ($(gtpc.east)+(\PL{\pipegap},0)$)
  {HPE model\\[2pt]
   {\scriptsize\textcolor{pipeGen}{\faFire}\; finetuned}};

\node[pimg,anchor=west] (kp2) at ($(hpe2.east)+(\PL{\pipegap},0)$)
  {\pipeimg{\imgRealKp}};
\node[plbl,above=0.5mm of kp2] (kp2lab) {3D joints};

\draw[parrow] (gtpc.east) -- (hpe2.west);
\draw[parrow] (hpe2.east) -- (kp2.west);

\coordinate (initY) at ($(boxTop)+(0,-5.5mm)$);
\coordinate (initL) at (hpe1.north |- initY);
\coordinate (initR) at (hpe2.north |- initY);

\draw[pdash]
  (hpe1.north) -- (initL)
  -- node[pcap,midway,above=0.8mm,fill=white,inner sep=1pt] {initialisation}
  (initR) -- (hpe2.north);

\coordinate (boxBottom) at ($(meshlab.south)-(0,\PL{\pipepadb})$);

\coordinate (genBoxL)     at ($(targetdblab.west)-(3mm,0)$);
\coordinate (genBoxR)     at ($(fm.east)+(\PL{\pipepad},0)$);

\coordinate (stageIBoxL)  at ($(gen.west)-(\PL{\padI},0)$);
\coordinate (stageIBoxR)  at ($(kp1.east)+(\PL{\padI},0)$);

\coordinate (stageIIBoxL) at ($(gtpc.west)-(\PL{\padII},0)$);
\coordinate (stageIIBoxR) at ($(kp2.east)+(\PL{\padII},0)$);

\begin{scope}[on background layer]
  \tikzset{panel/.style={draw=pipeRule,line width=1.2pt,rounded corners=3pt}}
  \draw[panel] (genBoxL     |- boxTop) rectangle (genBoxR     |- boxBottom);
  \draw[panel] (stageIBoxL  |- boxTop) rectangle (stageIBoxR  |- boxBottom);
  \draw[panel] (stageIIBoxL |- boxTop) rectangle (stageIIBoxR |- boxBottom);
\end{scope}

\coordinate (genBoxMid)     at ($(genBoxL)!0.5!(genBoxR)$);
\coordinate (stageIBoxMid)  at ($(stageIBoxL)!0.5!(stageIBoxR)$);
\coordinate (stageIIBoxMid) at ($(stageIIBoxL)!0.5!(stageIIBoxR)$);

\node[pstage,anchor=south] at ($(genBoxMid |- boxBottom)+(0,1mm)$)
  {\normalsize Data generation};

\node[pstage,anchor=south] at ($(stageIBoxMid |- boxBottom)+(0,1mm)$)
  {\normalsize Pretraining on synthetic data};

\node[pstage,anchor=south] at ($(stageIIBoxMid |- boxBottom)+(0,1mm)$)
  {\normalsize Finetuning on real data};

\end{tikzpicture}%
}

\caption{
\textbf{Our synthetic pretraining pipeline.} We sample local body poses $\theta$ from AMASS and bootstrap the remaining conditioning parameters, $\mathbf{o}$, $\beta$ as well as number of target points from the target dataset. These conditions are provided to our flow-matching model to generate novel human LiDAR observations. We then pretrain the HPE model on the generated samples and finetune it on real target dataset, yielding substantial performance gains.
}
\vspace{-3mm}
\label{fig:pipeline}
\end{figure*}
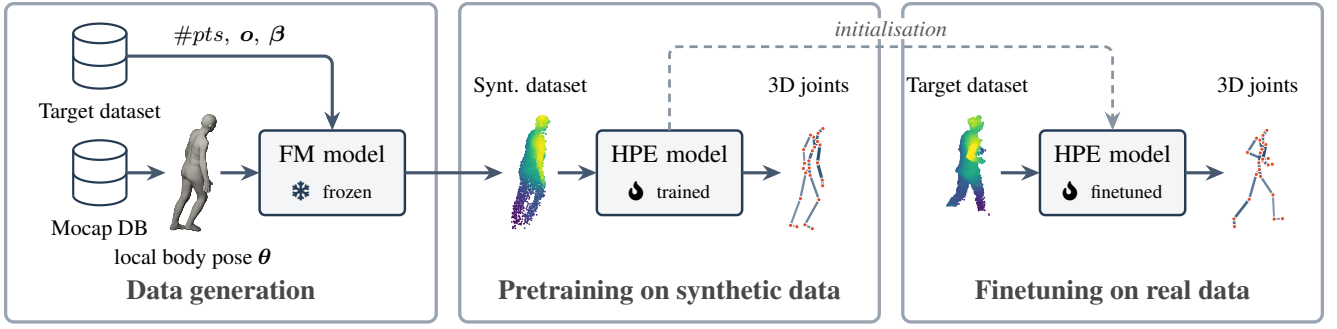

%% file: sec/4_exp.tex
\section{Experiments}
\label{sec:exp}
\input{tables/sloper_abl_merged}

\paragraph{Implementation details.}
To retain the characteristic of each LiDAR setup, we train separate models for different datasets. 
In the following experiments, we use a modified DiT-XS/4 configuration from~\cite{logen}, with 12 transformer blocks, a hidden dimension of 192, and three attention heads.

We train FM models for 200 epochs with a batch size of 128 and a learning rate of 1e-4. To handle variable-size point clouds, we pad each sample to the maximum number of points within the batch. For generation, we follow~\cite{ma_sit_2024} and solve the corresponding ordinary differential equation using the Euler solver with 50 sampling steps. 
All models are trained on a single NVIDIA H200 GPU and take roughly 15 and 3 hours to train on \sloper and \waymo, respectively.

\paragraph{Datasets.}
We conduct experiments on SLOPER4D \cite{sloper4d} and the Waymo Open Dataset (WOD) \cite{waymo}. SLOPER4D is an outdoor human motion-capture dataset containing synchronized LiDAR point clouds and RGB images, with ground-truth human meshes and 3D keypoints obtained using motion-capture devices. Since the dataset does not provide an official train/test split, we follow LiDAR-HMR~\cite{prn} and use 24,9k samples for training and 8,1k samples for testing.

For WOD, we use the human keypoint annotations provided with the v2.0 release, containing 8,1k annotated human instances for training and 1,8k for testing. 
As WOD does not provide ground-truth human meshes, we adopt the pseudo-mesh annotations from \cite{prn}.

\paragraph{Generative quality metrics.}
We evaluate generation quality on the validation set of each dataset using both sample-level and distribution-level metrics.

We measure the geometric similarity between a generated point cloud and the real observation with the same conditioning input using the Chamfer Distance (CD).

We assess distributional realism using the Fréchet PointNet Distance (FPD)~\cite{shu2019treegan}, adapted from FID~\cite{heusel2017fid}. We extract point-cloud representations from the PRN network \cite{prn}, using the features after its final multi-head self-attention layer and before the FCN head. We train a separate PRN for each dataset. We similarly adapt the Kernel Inception Distance (KID)~\cite{binkowski2018kid}, which is particularly suitable for smaller sample sets, and denote the resulting metric as Kernel PointNet Distance (KPD).

Finally, we compare the real and generated point-cloud distributions using coverage (COV)~\cite{achlioptas2018learningrepres} and nearest-neighbor accuracy (1-NNA)~\cite{yang2019pointflow,lopezpaz2018revisiting} following \cite{dit3d}, computed using CD. 
COV measures how much of the real distribution is covered by generated samples,  while 1-NNA measures how easily real and generated samples can be separated using nearest neighbors. A 1-NNA of $50\%$ indicates indistinguishable distributions.

\paragraph{Downstream evaluation.} 
To demonstrate the practical impact of our method, we use our trained FM models to generate new human LiDAR observations from poses in the AMASS database~\cite{amass}. Specifically, we use KIT and HumanEva sequences and filter them to \textit{walking}, \textit{standing} and \textit{running}, resulting in 47,494 unique body poses.
We sample the local body pose from AMASS, while bootstrapping the global body orientated location, shape and target point count from the target dataset, i.e., \sloper or \waymo. This places novel poses in realistic target-domain observation settings and allows the generator to produce realistic point densities and off-surface structures, rather than being limited to the returns obtained from ray casting a bare body mesh.
This provides a source of synthetic LiDAR samples with corresponding 3D keypoint annotations. 
We then pretrain a LiDAR-based 3D human pose estimator on the generated data, using the publicly available PRN~\cite{prn} as the downstream model. In a second stage, we finetune the pretrained PRN on real LiDAR data. We illustrate this pipeline in \cref{fig:pipeline}.
In both stages, we train PRN using 15 keypoints for 50 epochs using the Adam optimizer, with a batch size of 64 and a learning rate of 5e-4.
We evaluate downstream performance using MPJPE and PA-MPJPE, reported in centimeters; in the tables, PA denotes PA-MPJPE.

\input{tables/waymo_generative_results}
\input{tables/abl_amass_size}

\subsection{Experimental analysis}
\paragraph{Conditioning ablations.}
Since \sloper provides ground-truth human meshes, we conduct our ablation experiments on this dataset and report the generative quality metrics on the validation set. We begin with a simple setup without canonical coordinate transformation, conditioning only on the body pose $\theta$ while discarding the shape parameter $\beta$. As shown in \cref{tab:ablations}, both canonical transformation and shape conditioning improve performance when introduced separately and combining them yields further gains. Finally, replacing the angular pose with joint coordinates obtained through forward kinematics achieves the best performance. 

We also compare our generated samples against a raycasting baseline. For this baseline, we raycast the meshes from the corresponding validation sets using the respective LiDAR sensor specifications: 64 beams with a $20^\circ$ vertical FoV for WOD and 128 beams with a $45^\circ$ vertical FoV for SLOPER4D. As shown in \cref{tab:raycasting}, our generated point clouds achieve better generative quality.

In \cref{tab:waymo_gen}, we also report generative quality metrics for our final model on \waymo and compare it against the raycasting baseline. Consistent with the results on \sloper, our model achieves better overall generative quality.
We provide additional details regarding our method and ablations in the supplementary material.

\paragraph{Effect of pretraining with generated samples on LiDAR-based 3D HPE.}
We further evaluate the quality and practical utility of our generated samples on downstream LiDAR-based 3D human pose estimation. We first pretrain PRN on our generated human LiDAR data and then finetune the pretrained model on real human LiDAR instances. The results are summarized in \cref{tab:amass_size}. For both datasets, pretraining consistently outperforms training from scratch across different synthetic pretraining set sizes, and performance further improves as the amount of generated data increases. We also observe that the pretrained model alone, without real-data finetuning, achieves reasonable performance. 

\input{tables/abl_raycasting}

\paragraph{Comparison with raycasting pretraining.}
We further compare pretraining with our generated samples against raycasted samples using the same set of approximately 48k AMASS \cite{amass} meshes. For ray casting, we consider two settings. 
First, we ray cast the AMASS meshes using their original pose and body parameters, while simulating only the LiDAR beam pattern at the median range of the target-domain sensor.
Second, for a fair comparison with our generated data, we also bootstrap the body orientation, translation and shape parameters using real LiDAR samples before ray casting. 
In the tables, we denote the first setting as \emph{Rc — direct} and the second as \emph{Rc — bootstr.}
As shown in \cref{tab:raycasting}, pretraining with our generated samples consistently outperforms both raycasting variants in downstream 3D HPE.
\input{tables/sota_merged}

\subsection{Comparison to the-state-of-the-art}

We compare our pretraining pipeline using \ours-generated samples against existing LiDAR-based 3D HPE methods on the validation sets of \waymo and \sloper in \cref{tab:sota_all}, with qualitative results shown in \cref{fig:quals_merged} and \cref{fig:quals_amass}. For a fair comparison, we primarily consider methods that do not rely on additional large-scale real-world training datasets. In particular, DAPT~\cite{dapt} additionally uses LiDARHuman26M \cite{liLiDARCapLongrangeMarkerless2022}, which contains 184K LiDAR frames. PE-Proj~\cite{linHmPEARDatasetHuman2024} uses HmPEAR, which contains 250K frames with motion-capture-based 3D pose annotations.
For \sloper, some mesh-reconstruction methods do not explicitly report MPJPE. For these methods, we compute MPJPE from joints obtained from their predicted pose, shape and translation parameters to allow a consistent comparison.
Our approach achieves state-of-the-art performance on both datasets.

\subsection{Low-annotation regime}

Our approach can be particularly useful when LiDAR data is abundant but only a small fraction is annotated with 3D human keypoints. 
We investigate this on \waymo, where ground-truth meshes are unavailable and the body models are instead obtained using an off-the-shelf estimator \cite{prn}.
We use the full training set to learn our FM model, generate $\sim$48k human LiDAR samples from AMASS, and pretrain PRN on these samples. We then finetune PRN using only $1\%$ or $10\%$ of the labeled training data.
As shown in \cref{tab:low_annot_waymo}, this reduces the error by more than 50\% compared with training from scratch when using only $1\%$ and $10\%$ of the annotations. It also outperforms the raycasting baselines and GC-KPL~\cite{gckpl} in both low-annotation settings.

\input{tables/waymo_low_annot_regime}
\input{tables/low_data_regime}

\subsection{Low-data regime}
The other challenging setting where \ours can be particularly effective is the low-data regime, where only a limited amount of real LiDAR data is available, rather than merely limited annotations. 
To simulate this scenario, we restrict the entire pipeline to only $10\%$ and $20\%$ of the available real data. We first train the FM model on the corresponding low-data subset and use it to generate $\sim$48k synthetic human LiDAR samples from AMASS. We then pretrain the pose estimator on these generated samples and, in the final stage, finetune it using the same limited fraction of real data.
As shown in \cref{tab:low_data}, our pipeline substantially improves over training from scratch, reducing MPJPE and PA-MPJPE by approximately $30\%$ on SLOPER4D and $45\%$ on WOD. These results further demonstrate the effectiveness of our approach when real LiDAR data itself is scarce.

\input{figures/qualitative_merged}
\input{figures/qualitative_amass}

%% file: tables/sloper_abl_merged.tex
\begin{table*}[!t]
    \sisetup{
      round-mode            = places,
      round-precision       = 3,
      detect-all,
      table-align-text-post = false,
      input-symbols         = {*},
      text-font-command     = \underline,
    }
\caption{\textbf{Conditioning ablation study on \sloper using generative quality metrics}.
\underline{\emph{Canonical align.:}} \cmark~= canonical axis transformation applied; \xmark~= no canonical transformation.
\underline{\emph{Shape param.:}} \cmark~= shape parameters included in the conditioning; \xmark~= shape parameters excluded from the conditioning.
\underline{\emph{Cartesian joints:}} \cmark~= Cartesian joint coordinates with PointNet embeddings; \xmark~= 6D rotation coordinates with cyclic embeddings.}
    \label{tab:ablations}
    \centering
    \resizebox{0.8\textwidth}{!}{%
    \setlength{\tabcolsep}{4pt}
    \hspace*{-1ex}\begin{tabular}{c c c @{~} | @{~~} c c c c c }
        \toprule
         \textbf{Canonical align.} & \textbf{Shape param.} & \textbf{Cartesian joints} & \!\!\relax{\textbf{CD}}\,$\downarrow$ & \!\!\relax{\textbf{1-NNA}}\,$\downarrow$ & \relax{\textbf{COV}}\,$\uparrow$ & \!\!\relax{\textbf{FPD}}\,$\downarrow$ & \!\!\relax{\textbf{KPD}}\,\,$\downarrow$ \\
        \midrule
        \xmark & \xmark & \xmark & 0.02983 & 90.65 & 38.93 & 0.1857 & 0.01357 \\
        \cmark & \xmark & \xmark & 0.02836 & 90.46 & 40.21 & 0.1890 & 0.01689 \\
        \xmark & \cmark & \xmark & 0.02859 & 91.43 & 39.20 & 0.1725 & 0.01560 \\
        \cmark & \cmark & \xmark & 0.02718 & \underline{88.11} & \underline{43.99} & \B 0.1277 & 0.01273 \\
        \rowcolor{blue!10}\cmark & \cmark & \cmark &  \underline{0.02648} & \B 84.45 & \B 45.17 & 0.1508 & \B 0.01124 \\
        \midrule
         & Raycasting & &  \B 0.02470 & 95.45 & 37.77 & \underline{0.1390} & \underline{0.01177} \\
        \bottomrule
    \end{tabular}%
    }
    \vspace{-2mm}
\end{table*}

%% file: tables/waymo_generative_results.tex
\begin{table}[!t]
\caption{\textbf{Generative quality comparison on \waymo.}}
    \label{tab:waymo_gen}
    \centering
    \resizebox{0.49\textwidth}{!}{
    \setlength{\tabcolsep}{4pt}
    \hspace*{-1ex}\begin{tabular}{ c @{~} |@{~~} c  c c  c c }
        \toprule
         \textbf{Method} &  \textbf{CD} \,$\downarrow$ & \!\!\relax{\textbf{1-NNA}}\,$\downarrow$ & \relax{\textbf{COV}}\,$\uparrow$ & \textbf{FPD}\,$\downarrow$ & \!\!\textbf{KPD}\,\,$\downarrow$ 
         \\
        \midrule
    Raycasting & \textbf{0.0513} & 94.23  & 22.49 & 0.6051 & 0.07869 \\
    
\rowcolor{blue!10}\ours       & 0.0566 & \textbf{75.64} & \textbf{37.55} & \textbf{0.0961} & \textbf{0.00422} \\
 
        \bottomrule
    \end{tabular}
    }
\end{table}

%% file: tables/abl_amass_size.tex
\begin{table}[!t]
\captionof{table}{\textbf{Effect of synthetic pretraining on LiDAR-based 3D HPE.} $^\ast$ denotes results obtained by training on real data from scratch, without pretraining.
$\Delta$ values report the change in MPJPE relative to training from scratch without pretraining.}
\label{tab:amass_size}

\centering
\resizebox{0.49\textwidth}{!}{
\begin{tabular}{c c c c c c c}
\toprule

\multirow{3}{*}{\textbf{Dataset}} & \multirow{3}{*}{\shortstack{\textbf{Syn. pre.} \\ \textbf{data size} }}  & \multicolumn{2}{c}{\textbf{Pretraining}} & \multicolumn{2}{c}{\textbf{Finetuning}} & \multirow{3}{*}{$\mathbf{-\Delta\%}$}\\
\cmidrule(lr){3-4} \cmidrule(lr){5-6}
 &   & {\textbf{MPJPE}} & {\textbf{PA-}} & {\textbf{MPJPE}} & {\textbf{PA-}} & \\

\midrule
\parbox[t]{2mm}{\multirow{4}{*}{\rotatebox[origin=c]{90}{SLOPER4D}}} 

& \NA  & \NA & \NA &  6.02$^\ast$ & 4.58$^\ast$ & \NA \\
\cmidrule(lr){2-7}
& 8k    &   15.13  &   9.31   &   5.18   &   4.07 & 13.95 \\
& 24k   &   \underline{12.60}  &   \underline{7.63}   &   \underline{5.05}   &   \underline{4.02} & \underline{16.11} \\
&\cellcolor{blue!10}48k   &\cellcolor{blue!10}\B{11.09}
&\cellcolor{blue!10}\B{7.03}  &\cellcolor{blue!10}\B{5.04} &\cellcolor{blue!10}\B{3.97}  &\cellcolor{blue!10}\textbf{16.28} \\
\midrule
\parbox[t]{2mm}{\multirow{4}{*}{\rotatebox[origin=c]{90}{WOD}}} 

& \NA  & \NA & \NA & 6.74$^\ast$ & 5.09$^\ast$ & \NA \\
\cmidrule(lr){2-7}
& 8k    &   10.86  &   8.62   &   6.21   &   4.65  & 7.86 \\
& 24k   &   \B{10.63}  &   \underline{8.53}   &   {\underline{6.03}}   &   \underline{4.45} & \underline{10.53} \\
& \cellcolor{blue!10}48k   &\cellcolor{blue!10}\underline{10.79}  &\cellcolor{blue!10}\B{8.51}  &\cellcolor{blue!10}\B{5.85}   &\cellcolor{blue!10}\B{4.37} &\cellcolor{blue!10}\textbf{13.20} \\
\bottomrule
\end{tabular}
}
\vspace{-3mm}
\end{table}

%% file: tables/abl_raycasting.tex
\begin{table}[!t]
\captionof{table}{\textbf{Comparison of raycasted and \ours-generated samples using our pretraining pipeline on the downstream LiDAR-based 3D HPE task.} $\Delta$ values report the change in MPJPE relative to training from scratch without pretraining.
}
\label{tab:raycasting}

\centering
\resizebox{0.49\textwidth}{!}{
\begin{tabular}{c l c c c c c}
\toprule

\multirow{3}{*}{\textbf{Dataset}} & \multirow{3}{*}{\textbf{Method}}  & \multicolumn{2}{c}{\textbf{Pretraining}} & \multicolumn{2}{c}{\textbf{Finetuning}} & \multirow{3}{*}{$\mathbf{-\Delta\%}$}\\
\cmidrule(lr){3-4} \cmidrule(lr){5-6}
 &   & {\textbf{MPJPE}} & {\textbf{PA-}} & {\textbf{MPJPE}} & {\textbf{PA-}} & \\

\midrule
\parbox[t]{2mm}{\multirow{3}{*}{\rotatebox[origin=c]{90}{\footnotesize SLOPER4D}}} 

&\cellcolor{blue!10}\ours   
&\cellcolor{blue!10}\underline{11.09}  &\cellcolor{blue!10}\B{7.03}   
&\cellcolor{blue!10}\B{5.04} 
&\cellcolor{blue!10}\B{3.97} 
&\cellcolor{blue!10}\B{16.28}\\
\cmidrule(lr){2-7}

& Rc — bootstr.   &  \B 11.03 & \underline{7.26}   &   5.47 & 4.26  & 9.14 \\
& Rc — direct    &  21.95 & 9.73    &   \underline{5.27} & \underline{4.16} & \underline{12.45} \\ 
\midrule

\parbox[t]{2mm}{\multirow{3}{*}{\rotatebox[origin=c]{90}{\footnotesize WOD}}} 
&\cellcolor{blue!10}\ours  
&\cellcolor{blue!10}\B{10.79}  
&\cellcolor{blue!10}\B{8.51}   
&\cellcolor{blue!10}\B{5.85} 
&\cellcolor{blue!10}\B{4.37} 
&\cellcolor{blue!10}\B{13.20} \\
\cmidrule(lr){2-7}
& Rc — bootstr.    &  15.54 & 10.39    &   \underline{5.98} & \underline{4.42} & \underline{11.28}  \\
& Rc — direct &  \underline{15.11}  &   \underline{10.04}    &   6.35  & 4.71  & 5.79 \\ 
\bottomrule
\end{tabular}
}
\vspace{-2mm}
\end{table}

%% file: tables/sota_merged.tex
\begin{table}[htbp]
\centering

\caption{\textbf{SOTA comparison for LiDAR-based 3D HPE.} \underline{\emph{Bench.:}} \cmark~= the benchmark data is the only real-data source used in the pipeline; \xmark~= training includes extra large-scale real data.  \underline{\emph{LiDAR:}} \cmark~= input is only LiDAR point clouds; \xmark~= input includes images. }
\label{tab:sota_all}
\vspace{-0.7em}
\begin{subtable}{\linewidth}
  \centering
  \caption{Comparison on the \waymo~\cite{waymo} validation split.}
  \label{tab:benchmark_hpe_wod_val}
  \resizebox{\linewidth}{!}{%
  \begin{tabular}{lcccc}
  \toprule
  \textbf{Method} & \textbf{Reference} & \textbf{Bench.} & \textbf{LiDAR} & \textbf{MPJPE} \\
  \midrule
  DAPT & \cite{dapt}\rf{aaai}{25} & \xmark & \cmark & \textbf{5.16} \\
  \midrule
  WS-HPE & \cite{zhengMultimodal3DHuman2022}\rf{cvprw}{22} & \cmark & \xmark & 10.32 \\
  WS-Fusion & \cite{bauerWeaklySupervisedMultiModal2023}\rf{ieee-iv}{23} & \cmark & \xmark & 8.58 \\
  HUM3DIL & \cite{zanfirHUM3DILSemisupervisedMultimodal2022}\rf{corl}{22} & \cmark & \xmark & 6.72 \\
  GC-KPL & \cite{gckpl}\rf{cvpr}{23} & \cmark & \cmark & 10.10 \\
  VoxelKP & \cite{shiVoxelKPVoxelbasedNetwork2023}\rf{iccv}{25} & \cmark & \cmark & 8.87 \\
  PRN & \cite{prn}\rf{ieee-tmm}{25} & \cmark & \cmark & 6.78 \\
  LPFormer & \cite{yeLPFormerLiDARPose2024}\rf{icra}{24} & \cmark & \cmark & 6.16 \\
  \rowcolor{blue!10}
  \ours & ours & \cmark & \cmark & \textbf{5.85} \\
  \bottomrule
  \end{tabular}%
  }
\end{subtable}

\vspace{0.5em}

\begin{subtable}{\linewidth}
  \centering
  \caption{Comparison on the \sloper~\cite{sloper4d} validation split.}
  \label{tab:benchmark_hmr_sloper_val}
  \resizebox{\linewidth}{!}{%
  \begin{tabular}{lcccc}
  \toprule
  \textbf{Method} & \textbf{Reference} & \textbf{Bench.} & \textbf{LiDAR} & \textbf{MPJPE} \\
  \midrule
  PE-Proj & \cite{linHmPEARDatasetHuman2024}\rf{acm mm}{24} & \xmark & \xmark & 5.55 \\
  DAPT & \cite{dapt}\rf{aaai}{25} & \xmark & \cmark & \textbf{2.80} \\
  \midrule
  LiDARCap & \cite{liLiDARCapLongrangeMarkerless2022}\rf{cvpr}{22} & \cmark & \cmark & 8.61 \\
  LIP & \cite{renLiDARaidInertialPoser2023}\rf{ieee-tvcg}{24} & \cmark & \xmark & 9.48 \\
  LiveHPS & \cite{renLiveHPSLiDARbasedScenelevel2024}\rf{cvpr}{24} & \cmark & \cmark & 7.77 \\
  LiveHPS++ & \cite{renLiveHPSRobustCoherent2024}\rf{eccv}{24} & \cmark & \cmark & 7.70 \\
  NE-3D-HPE & \cite{zhangNeighborhoodEnhanced3DHuman2024}\rf{aaai}{24} & \cmark & \cmark & 9.68 \\
  LiDAR-HMR & \cite{prn}\rf{ieee-tmm}{25} & \cmark & \cmark & 5.10 \\
  PRN & \cite{prn}\rf{ieee-tmm}{25} & \cmark & \cmark & 6.02 \\
  ReMP & \cite{jangReMPReusableMotion2025}\rf{wacv}{25} & \cmark & \cmark & 21.66 \\
  \rowcolor{blue!10}
  \ours & ours & \cmark & \cmark & \textbf{5.04} \\
  \bottomrule
  \end{tabular}%
  }
\end{subtable}
\vspace{-5mm}
\end{table}

%% file: tables/waymo_low_annot_regime.tex
\begin{table}[!t]
\captionof{table}{\textbf{Results in the low-annotation regime on the \waymo validation set.} Zero-shot denotes evaluation directly using models pretrained on synthetic data, without finetuning on real data. \emph{Pt.} denotes \emph{pretraining.}
}
\label{tab:low_annot_waymo}

\centering
\resizebox{0.49\textwidth}{!}{
\begin{tabular}{l c c c c c c c c}
\toprule
\multirow{3}{*}{\textbf{Method}} & \multirow{3}{*}{\textbf{Pt.}}  & \multicolumn{2}{c}{\textbf{Zero-shot (0\%)}} & \multicolumn{2}{c}{\textbf{1 \%}}  & \multicolumn{2}{c}{\textbf{10 \%}} \\

\cmidrule(lr){3-4} \cmidrule(lr){5-6} \cmidrule(lr){7-8}
& & {\textbf{MPJPE}} & {\textbf{PA-}}  & {\textbf{MPJPE}} & {\textbf{PA-}} & {\textbf{MPJPE}} & {\textbf{PA-}}  \\ 

\midrule

Random init. & \xmark   & \na & \na & 18.01 & 14.19  & 13.74 & 10.60  \\

Rc — bootstr. & \cmark   & 15.54 & 10.39 & \underline{9.01} & \underline{7.15}  & \underline{7.22} & \underline{5.45}  \\

Rc — direct & \cmark   & \underline{15.11}  &   \underline{10.04} & 10.89 & 8.37 & 7.98 & 6.16  \\

GC-KPL \cite{gckpl} & \cmark & 17.70 & {-} & 18.52 & {-} & 15.10 & {-}  \\

\rowcolor{blue!10}\ours & \cmark & \B 10.79 & \B 8.51  & \B 8.43 & \B 6.48 & \B 6.84 & \B 5.18    \\

\bottomrule
\end{tabular}
}
\vspace{-3mm}
\end{table}

%% file: tables/low_data_regime.tex
\begin{table}[!t]
\captionof{table}{\textbf{Results in the low-data regime on the \waymo and \sloper validation sets using 10\% and 20\% of the real training data.} 
Pt: pretraining; Ft: finetuning.}
\label{tab:low_data}

\centering
\resizebox{0.49\textwidth}{!}{
\begin{tabular}{l c c S S c S S c}
\toprule
\multirow{3}{*}{\textbf{Fraction}} & \multirow{3}{*}{\textbf{Pt.}} & \multirow{3}{*}{\textbf{Ft.}} & \multicolumn{2}{c}{\textbf{SLOPER4D}} & \multirow{3}{*}{$\mathbf{-\Delta\%}$} & \multicolumn{2}{c}{\textbf{WOD}} & \multirow{3}{*}{$\mathbf{-\Delta\%}$}\\
\cmidrule(lr){4-5} \cmidrule(lr){7-8}
 & &  & {\textbf{MPJPE}} & {\textbf{PA-}} & & {\textbf{MPJPE}} & {\textbf{PA-}}\\
\midrule

\multirow{3}{*}{{10\%}} 
& \xmark  & \cmark &  9.292 & 6.618 & &  13.273 & 10.267 \\
& \cmark  & \xmark & 16.398 & 9.376 & & 14.072 & 8.857 \\
& \cmark  & \cmark &  \B 6.464 & \B 4.863 & \B 30.46 & \B 7.568 & \B 5.614 & \B 42.95 \\
\midrule
\multirow{3}{*}{{20\%}} 
& \xmark & \cmark   &  7.906 & 5.603 & & 10.526 & 8.281 \\
& \cmark  & \xmark & 13.825 & 8.646 & & 13.775 & 8.773 \\
& \cmark & \cmark  &  \B 5.872 & \B 4.530 & \B 25.79 & \B 6.693 & \B 5.012 & \B 36.47\\
\bottomrule
\end{tabular}
}
\vspace{-5mm}
\end{table}

%% file: figures/qualitative_merged.tex
\providecommand{\qbase}{figures/teaser_viridis}

\ifdefined\qh\else\newlength{\qh}\fi
\setlength{\qh}{2.12cm}

\ifdefined\qw\else\newlength{\qw}\fi
\setlength{\qw}{0.72\qh}

\ifdefined\qv\else\newlength{\qv}\fi
\setlength{\qv}{4pt}

\providecommand{\qc}[1]{%
  \makebox[\qw][c]{%
    \includegraphics[height=\qh]{\qbase/#1}%
  }%
}

\providecommand{\qhd}[1]{%
  \makebox[\qw][c]{\footnotesize #1}%
}

\providecommand{\qn}[1]{%
  \makebox[\qw][c]{\scriptsize #1}%
}

\providecommand{\qlab}[1]{%
  \rotatebox[origin=c]{90}{\footnotesize\textsc{#1}}%
}

%
\providecommand{\qpair}[8]{%
  %
  \qc{#1/rgb.png} &
  \qc{#1/gt_mesh.pdf} &
  \qc{#1/gt_pc_#2.pdf} &
  \qc{#1/ray_cast_#3.pdf} &
  \qc{#1/ours_#4.pdf} &
  %
  \qc{#5/rgb.png} &
  \qc{#5/gt_mesh.pdf} &
  \qc{#5/gt_pc_#6.pdf} &
  \qc{#5/ray_cast_#7.pdf} &
  \qc{#5/ours_#8.pdf}
  \tabularnewline[1pt]
  %
    \multicolumn{2}{r}{\scriptsize \# pts\hspace{3pt}} &
    \qn{#2} &
    \qn{#3} &
    \qn{#4} &
    \multicolumn{2}{r}{\scriptsize \# pts\hspace{3pt}} &
    \qn{#6} &
    \qn{#7} &
    \qn{#8}
  \tabularnewline[\qv]
}

\begin{figure*}[t]
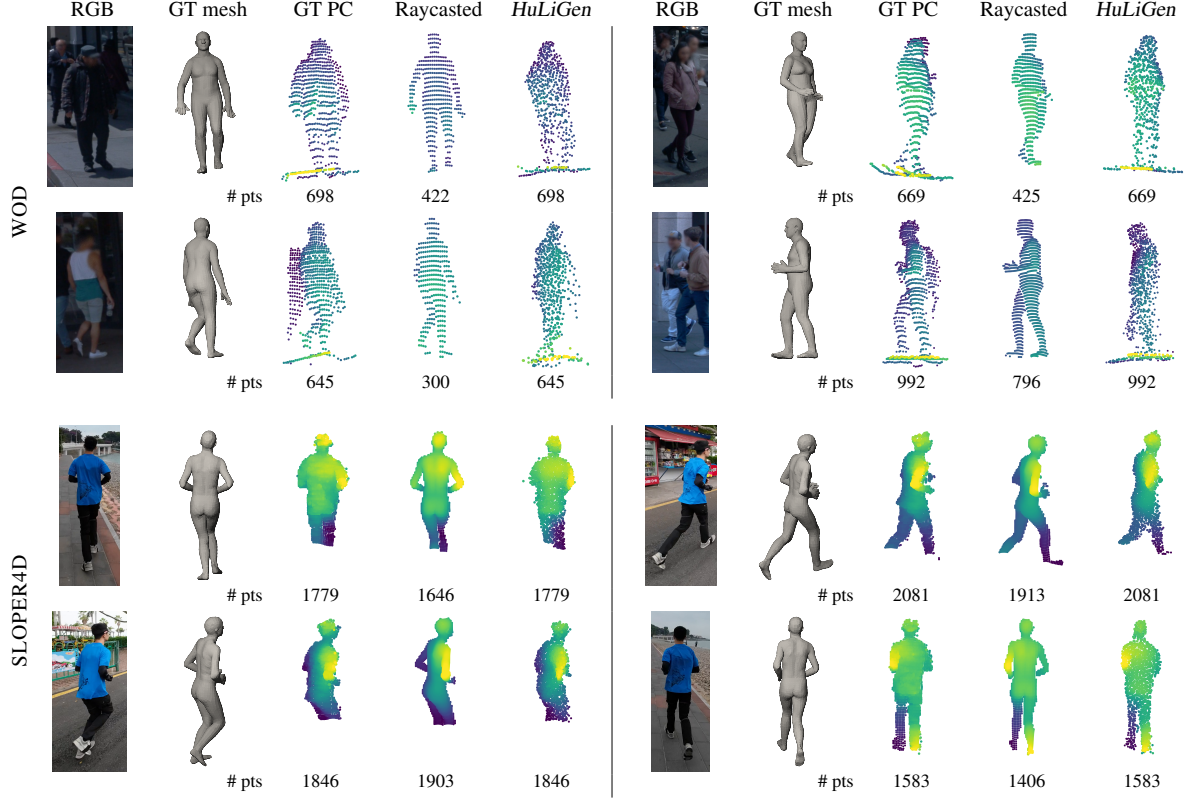

\centering

\setlength{\tabcolsep}{0pt}
\renewcommand{\arraystretch}{0}

\begin{tabular}{@{}c@{\hspace{2pt}}c@{}}

&
\begin{tabular}{
  @{}
  ccccc
  @{\hspace{3pt}}
  ccccc
  @{}
}
  \qhd{RGB} &
  \qhd{GT mesh} &
  \qhd{GT PC} &
  \qhd{Raycasted} &
  \qhd{\ours} &
  \qhd{RGB} &
  \qhd{GT mesh} &
  \qhd{GT PC} &
  \qhd{Raycasted} &
  \qhd{\ours}
\end{tabular}
\tabularnewline[3pt]

\qlab{WOD} &
\begin{tabular}{
  @{}
  cc
  c
  cc
  @{\hspace{1pt}\vrule width 0.3pt\hspace{1pt}}
  @{\hspace{3pt}}
  cc
  c
  cc
  @{}
}

  \qpair
    {waymo_val_2551868399007287341_3100_000_3120_000_q_PAmaMLwQ3zf6cXBDDIag_x-1558465505164513}
    {698}{422}{698}
    {waymo_val_4764167778917495793_860_000_880_000_C8XJW3cxR3DsnPQZX0ILIg_x-1557886551147548}
    {669}{425}{669}

  \qpair
    {waymo_val_7493781117404461396_2140_000_2160_000_V6TdgYsnJmV016TzDDcYAg_x-1559172284850559}
    {645}{300}{645}
    {waymo_val_8079607115087394458_1240_000_1260_000_Mb6Z3Qd6jBsk5IFBnvPMLQ_x-1557265259162469}
    {992}{796}{992}

\end{tabular}
\tabularnewline[4pt]

\tabularnewline[10pt]

\qlab{SLOPER4D} &
\begin{tabular}{
  @{}
  cc
  c
  cc
  @{\hspace{1pt}\vrule width 0.3pt\hspace{1pt}}
  @{\hspace{3pt}}
  cc
  c
  cc
  @{}
}

\qpair
    {sloper4d_val_seq009_running_002-1054_771}
    {1779}{1646}{1779}
    {sloper4d_val_seq009_running_002-1218_004}
    {2081}{1913}{2081}

  \qpair
    {sloper4d_val_seq009_running_002-1225_304}
    {1846}{1903}{1846}
    {sloper4d_val_seq009_running_002-952_581}
    {1583}{1406}{1583}

\end{tabular}

\end{tabular}

\caption{
\textbf{Qualitative results on the \waymo \cite{waymo} and \sloper \cite{sloper4d} validation sets.} 
For each sample, we show the camera crop, body-model surface, real LiDAR observation, raycasted result and our generation, with point counts below. Camera images are for visualization only, and all panels share the same viewpoint and scale. Ray casting stays on the bare body surface, while real and generated observations capture off-surface structure from clothing and hair.
}

\label{fig:quals_merged}

\end{figure*}

%% file: figures/qualitative_amass.tex
\providecommand{\qbase}{figures/teaser_viridis}

\ifdefined\qh\else\newlength{\qh}\fi
\setlength{\qh}{2.3cm}

\ifdefined\qw\else\newlength{\qw}\fi
\setlength{\qw}{0.76\qh}

\ifdefined\qv\else\newlength{\qv}\fi
\setlength{\qv}{4pt}

\providecommand{\qc}[1]{%
  \makebox[\qw][c]{%
    \includegraphics[height=\qh]{\qbase/#1}%
  }%
}

\providecommand{\qhd}[1]{%
  \makebox[\qw][c]{\footnotesize #1}%
}

\providecommand{\qn}[1]{%
  \makebox[\qw][c]{\scriptsize #1}%
}

\providecommand{\qlab}[1]{%
  \rotatebox[origin=c]{90}{\footnotesize\textsc{#1}}%
}

\providecommand{\qtriplesloper}[9]{%
  %
  \qc{#1/gt_mesh.pdf} &
  \qc{#1/ray_cast_#2.pdf} &
  \qc{#1/ours_#3.pdf} &
  %
  \qc{#4/gt_mesh.pdf} &
  \qc{#4/ray_cast_#5.pdf} &
  \qc{#4/ours_#6.pdf} &
  %
  \qc{#7/gt_mesh.pdf} &
  \qc{#7/ray_cast_#8.pdf} &
  \qc{#7/ours_#9.pdf}
  \tabularnewline[2pt]
  %
  \multicolumn{1}{r}{\scriptsize \# pts\hspace{3pt}} &
  \qn{#2} &
  \qn{#3} &
  \multicolumn{1}{r}{\scriptsize \# pts\hspace{3pt}} &
  \qn{#5} &
  \qn{#6} &
  \multicolumn{1}{r}{\scriptsize \# pts\hspace{3pt}}  &
  \qn{#8} &
  \qn{#9}
  \tabularnewline[\qv]
}

\providecommand{\qtriplewod}[9]{%
  %
  \qc{#1/gt_mesh_wod.pdf} &
  \qc{#1/ray_cast_wod_#2.pdf} &
  \qc{#1/ours_wod_#3.pdf} &
  %
  \qc{#4/gt_mesh_wod.pdf} &
  \qc{#4/ray_cast_wod_#5.pdf} &
  \qc{#4/ours_wod_#6.pdf} &
  %
  \qc{#7/gt_mesh_wod.pdf} &
  \qc{#7/ray_cast_wod_#8.pdf} &
  \qc{#7/ours_wod_#9.pdf}
  \tabularnewline[2pt]
  %
  \multicolumn{1}{r}{\scriptsize \# pts\hspace{3pt}} &
  \qn{#2} &
  \qn{#3} &
  \multicolumn{1}{r}{\scriptsize \# pts\hspace{3pt}} &
  \qn{#5} &
  \qn{#6} &
  \multicolumn{1}{r}{\scriptsize \# pts\hspace{3pt}}  &
  \qn{#8} &
  \qn{#9}
  \tabularnewline[\qv]
}

\begin{figure*}[t]
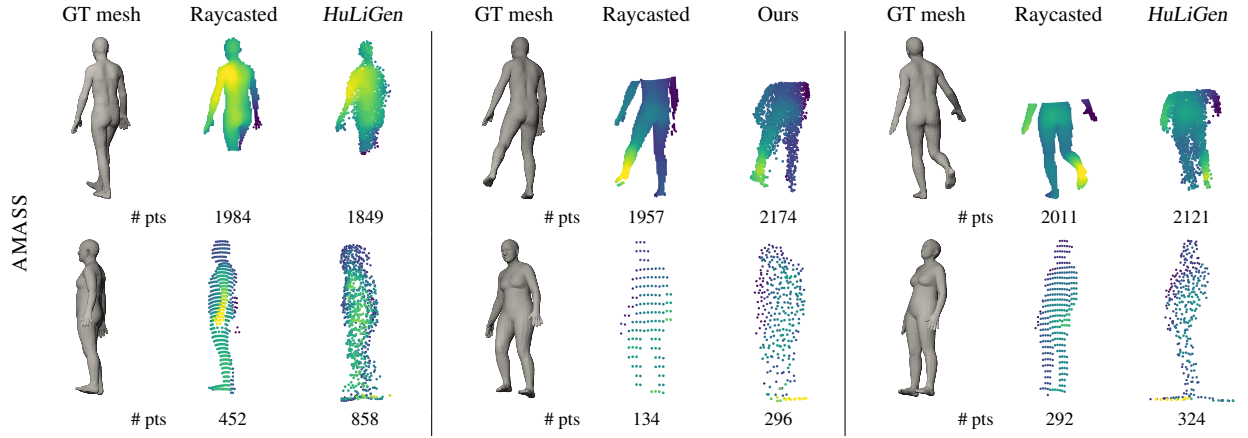

\centering

\setlength{\tabcolsep}{0pt}
\renewcommand{\arraystretch}{0}

\begin{tabular}{@{}c@{\hspace{3pt}}c@{}}

&
\begin{tabular}{
  @{}
  ccc
  @{\hspace{6pt}}
  ccc
  @{\hspace{6pt}}
  ccc
  @{}
}
  \qhd{GT mesh} &
  \qhd{Raycasted} &
  \qhd{\ours} &
  \qhd{GT mesh} &
  \qhd{Raycasted} &
  \qhd{Ours} &
  \qhd{GT mesh} &
  \qhd{Raycasted} &
  \qhd{\ours}
\end{tabular}
\tabularnewline[3pt]

\qlab{AMASS} &
\begin{tabular}{
  @{}
  ccc|
  @{\hspace{6pt}}
  ccc|
  @{\hspace{6pt}}
  ccc
  @{}
}

\qtriplesloper
  {amass_train_291-push_recovery_stand_left04_stageii-169}{1984}{1849}
  {amass_train_291-push_recovery_stand_right03_stageii-227}{1957}{2174}
  {amass_train_3-walking_fast05_stageii-138}{2011}{2121}

\qtriplewod
  {amass_train_3-walking_run02_stageii-491}{452}{858}
  {amass_train_314-walking_slow08_stageii-79}{134}{296}
  {amass_train_425-walking_fast05_stageii-347}{292}{324}

\end{tabular}

\end{tabular}

\caption{
\textbf{Qualitative results on AMASS.} 
AMASS provides pose and shape but no sensor data, so we use the body surface as reference. For each sample, we show the body surface, raycast result and our generation, with point counts below. Ray casting remains on the bare body surface, yielding a median surface distance of 0 cm on both datasets. In contrast, our generations reach 1.41 cm on SLOPER4D (top row), close to 1.17 cm for real observations, and 4.5 cm on WOD compared with 3.0 cm for real data. We attribute the larger WOD (bottom row) gap mainly to ground-level noise.
}

\label{fig:quals_amass}
\vspace{1cm}
\end{figure*}

%% file: sec/5_conclusion.tex
\section{Conclusion}
\label{sec:conc}
In this paper, we introduced \ours, a point-transformer-based flow-matching model for human LiDAR generation conditioned on parametric body models. \ours produces more realistic observations than standard raycasting and allows large pose repositories to be converted into synthetic LiDAR data. We show that pretraining HPE on these samples significantly improves performance, with particularly strong gains in low-data and low-annotation regimes.

%% file: sec/X_suppl.tex
\clearpage
\setcounter{page}{1}
\maketitlesupplementary

\input{tables/abl_params}

This appendix supplements the main paper with additional details on our method and ablation experiments.

\section{Complementary details on conditioning and embedding}

Here, we provide additional details on the conditioning formulations explored in our ablation study. In particular, we describe how our Flow Matching framework is instantiated with the SMPL~\cite{loper2023smpl} body model and explain the different conditioning and embedding strategies considered.

\paragraph{Data format.}
We use the point cloud $\mathbf{P}\in\mathbb{R}^{N\times3}$, transformed into the sensor frame, together with the corresponding SMPL annotations: shape $\beta\in\mathbb{R}^{10}$, full local body pose $\theta\in\mathbb{R}^{72}$ represented by 24 axis-angle triplets, and translation $t_l$. The global orientation $\theta_0$ and translation $t_l$ are also expressed in the sensor frame. All ablation experiments in Table~1 of the main paper use the same $\mathbf{P}$, $t_l$, and $\theta$; they differ only in whether the shape parameter $\beta$ is included and in how the pose $\theta$ is provided to the network as well as canonicalisation. 
We provide an overview of the ablation configurations in Table~\ref{tab:ablations_params}, detailing the representations, conditioning choices and dimentionality
used in each setting.

\paragraph{Canonicalisation.}

For canonicalisation, our model uses the lateral axis of the SMPL root joint. We first convert the root axis-angle representation $\boldsymbol{\theta}_0$ into a rotation matrix,
 
\begin{equation} \mathbf{R}_0=\mathcal{R}(\boldsymbol\theta_0)\in SO(3) \end{equation}

expressed in the sensor frame. The columns of $\mathbf{R}_0$ correspond to the transformed SMPL canonical body axes: $\mathbf{R}_0\hat e_1$ denotes the \emph{lateral} axis, $\mathbf{R}_0\hat e_2$ the \emph{upward} axis, and $\mathbf{R}_0\hat e_3$ the \emph{forward} axis. We project the lateral axis onto the horizontal plane and compute its azimuth $\psi$, which defines the canonicalising rotation:

\begin{equation}
\psi \;=\; \operatorname{atan2}\!\big([\mathbf{R}_0]_{10},\,[\mathbf{R}_0]_{00}\big),
\label{eq:psi}
\end{equation}

and $\mathbf{R}_z(-\psi)$ is applied consistently to all geometric quantities.

\begin{algorithm}[h]
\caption{Canonicalisation}
\label{alg:canon}
\begin{algorithmic}[1]
\Require points $\mathbf{P}$, pose $\theta$, shape $\beta$, translation $t_l$
\State ${t_l}\gets$ body centre; $(\phi,\rho,z)\gets\mathrm{cyl}(t)$
\State $\mathbf{R}_0\gets\mathcal{R}(\boldsymbol\theta_0)$
\State $\psi\gets\operatorname{atan2}([\mathbf{R}_0]_{10},[\mathbf{R}_0]_{00})$ 
\State $\mathbf{P}\gets\mathbf{R}_z(-\psi)\cdot(\mathbf{P}-t_l)$ 
\State $\boldsymbol\theta_0\gets\mathrm{rotvec}\big(\mathbf{R}_z(-\psi)\cdot\mathbf{R}_0\big)$ 
\State $\phi_{\mathrm{rel}}\gets\mathrm{wrap}(\phi-\psi)$ 
\State \Return $\mathbf{P}$, $\boldsymbol\theta$, $\boldsymbol\beta$, $\mathbf{o}=(\phi_{\mathrm{rel}},\rho,z)$
\end{algorithmic}
\end{algorithm}

Using the projected lateral axis rather than the forward axis avoids degenerate cases in which the forward direction approaches the vertical axis, making its projection onto the horizontal plane, and consequently its azimuth, ill-defined. Examples from \sloper \cite{sloper4d} are shown in \cref{fig:root_axes_poses2}, where crouched poses illustrate this issue and motivate our choice. We present our canonicalization algorithm in \cref{alg:canon}.

For the ablations without canonicalisation (corresponding to rows 1 and 3 of Table~1 in the main paper), we instead express the translation $t_l$ in cylindrical coordinates. Its azimuth $\phi$ is encoded as $(\sin\phi,\cos\phi)$ and provided to the network together with $\log\rho$ and $z$. $\mathbf{o}$ thus becomes $(\sin\phi,\cos\phi,\log\rho,z)$.

\paragraph{Local body pose representations.}
Our model applies the SMPL forward kinematics (FK) function to the input axis-angle poses $\theta$ to obtain the corresponding 3D joint positions, which are then embedded using the joint-position encoder described in Sec.~3.3 of the main paper. 
As part of our ablation study, we also consider an alternative pose representation in which the axis-angle joint rotations are converted to 6D rotations with cyclic embedding, together with a learned positional encoding analogous to that used in the FK-based formulation.

%% file: tables/abl_params.tex
\begin{table*}[!t]
\caption{\textbf{Conditioning and embedding details}.
\underline{\emph{Canonical align.:}} \cmark~= canonical axis transformation applied; \xmark~= no canonical transformation.
\underline{\emph{Shape param.:}} \cmark~= shape parameters included in the conditioning; \xmark~= shape parameters excluded from the conditioning.
\underline{\emph{Cartesian joints:}} \cmark~= Cartesian joint coordinates with PointNet embeddings; \xmark~= 6D rotation coordinates with cyclic embeddings. 
\emph{abs. cyl.} denotes absolute cylindrical coordinates and \emph{canon.} denotes canonicalized coordinates. 
The third column group reports the dimensionality of each conditioning representation before embedding;
while the final column gives the total conditioning-token count used by each variant.
}
    \label{tab:ablations_params}
    \centering
    \resizebox{\linewidth}{!}{%
    \setlength{\tabcolsep}{4pt}
    \hspace*{-1ex}\begin{tabular}{c c c @{~} | @{~~} c c @{~} | @{~~} c c c @{~} | @{~~} c}
        \toprule
         \textbf{Canonical align.} & \textbf{Shape param.} & \textbf{Cartesian joints} & $\mathbf{o}_{rep}$ & $\theta_{rep}$ & $o_{emb}$ & $\beta_{emb}$ & $\theta_{emb}$ & token count\\
        \midrule
        \xmark & \xmark & \xmark & abs. cyl. & 24 $\times$ 6D rotation & 4 & \na & 144 & 28 \\
        \cmark & \xmark & \xmark & canon. & 24 $\times$ 6D rotation & 3 & \na & 144 & 27 \\
        \xmark & \cmark & \xmark & abs. cyl. & 24 $\times$ 6D rotation & 4 & 10 & 144 & 38 \\
        \cmark & \cmark & \xmark & canon. & 24 $\times$ 6D rotation & 3 & 10 & 144 & 37 \\
        \cmark & \cmark & \cmark & canon. & 24 $\times$ 3D position & 3 & 10 & 72 & 37 \\
        \bottomrule
    \end{tabular}%
    }
    \vspace{-2mm}
\end{table*}